\documentclass{article}
\usepackage{ijcai18}

\usepackage{times}
\usepackage{xcolor}
\usepackage{soul}
\usepackage[utf8]{inputenc}
\usepackage[small]{caption}

\usepackage{graphicx} 
\usepackage[boxed]{algorithm2e}
\usepackage{mathptmx} 
\usepackage{amsmath} 
\usepackage{amsfonts}
\usepackage{url} 

\title{Fast Model Identification via Physics Engines for Data-Efficient Policy Search}

\author{
Shaojun Zhu, 
Andrew Kimmel,
Kostas E. Bekris, 
Abdeslam Boularias
\\ 
Department of Computer Science, Rutgers University, New Jersey, USA \\
\{shaojun.zhu, ask139, kostas.bekris, ab1544\} @cs.rutgers.edu
}

\makeatletter
\newcommand{\removelatexerror}{\let\@latex@error\@gobble}
\makeatother

\begin{document}

\maketitle

\begin{abstract}
This paper presents a method for identifying mechanical parameters of
robots or objects, such as their mass and friction coefficients. Key
features are the use of off-the-shelf physics engines and the
adaptation of a Bayesian optimization technique towards minimizing the
number of real-world experiments needed for model-based reinforcement
learning. The proposed framework reproduces in a physics engine
experiments performed on a real robot and optimizes the model's
mechanical parameters so as to match real-world trajectories. The
optimized model is then used for learning a policy in simulation,
before real-world deployment. It is well understood, however, that it
is hard to exactly reproduce real trajectories in simulation.
Moreover, a near-optimal policy can be frequently found with an
imperfect model. Therefore, this work proposes a strategy for
identifying a model that is just good enough to approximate the value
of a locally optimal policy with a certain confidence, instead of
wasting effort on identifying the most accurate model. Evaluations,
performed both in simulation and on a real robotic manipulation task,
indicate that the proposed strategy results in an overall
time-efficient, integrated model identification and learning solution,
which significantly improves the data-efficiency of existing policy
search algorithms.
\end{abstract}

\section{Introduction}

\begin{figure}[!t]
  \centering
  \includegraphics[width=0.45\textwidth]{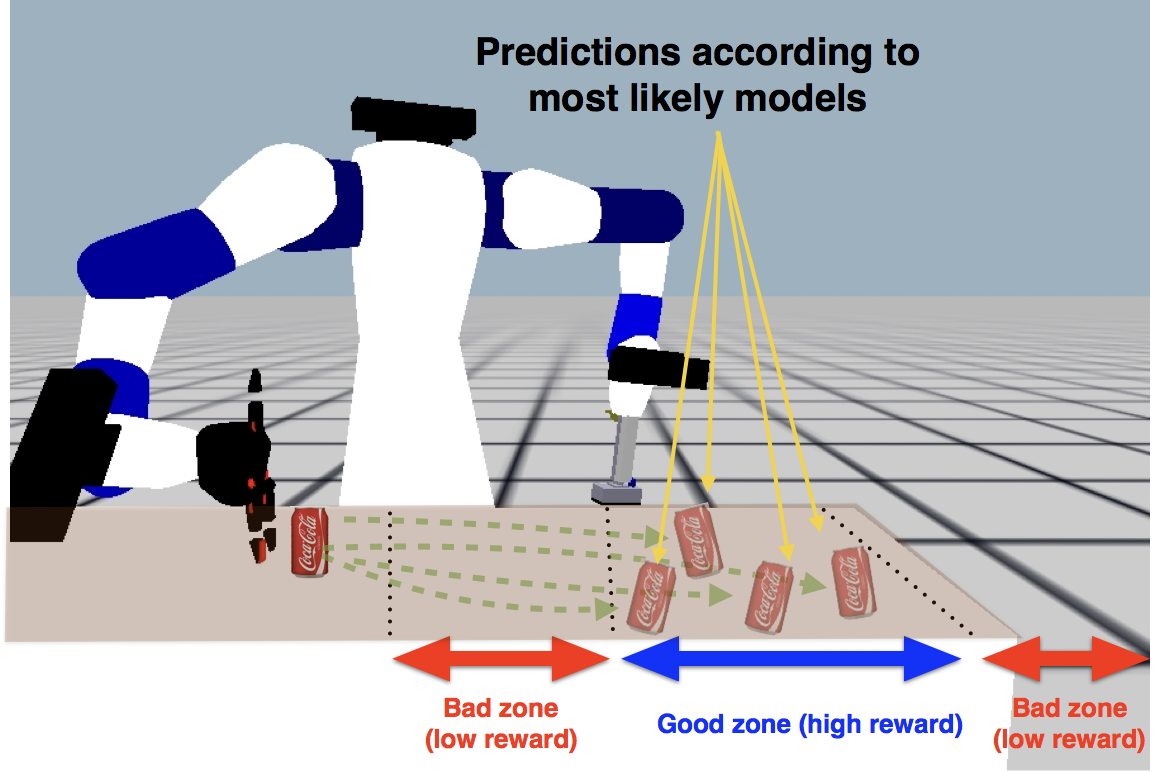}
\caption{\small Example of a stopping condition for model
  identification: if all high-probability models predict a high reward
  for a given action, there is no point in singling out the most accurate
  model.}
\end{figure}

Reinforcement learning (RL) typically requires a lot of training data
before it can provide useful skills in robotics. Learning in
simulation can help reduce the dependence on real-world data but the
learned policy may not work due to model inaccuracies, also known as
the ``reality gap''.  This paper presents an approach for model
identification by exploiting the availability of off-the-shelf physics
engines~\cite{ErezTT15} for simulating robot and object dynamics in a
given scene. One objective is to achieve data-efficiency for RL by
reducing the need for many real-world robot experiments by providing
better-simulated models.  It also aims to achieve time efficiency in
the context of model identification by reducing the computational
effort of this process.


The accuracy of a physics engine depends on several factors. The model assumed by
the engine, such as Coulomb's law of friction, may be limited or the
numerical algorithm for solving the differential equations of motion
may introduce errors.  Moreover, the  robot's and the objects'
mechanical parameters, such as mass, friction, and elasticity, may be
inaccurate. This work focuses on this last factor and proposes a
method for identifying mechanical parameters used in physical
simulations so as to assist in learning a robotic task.

Given initial real-world trajectories, potentially generated by
randomly initializing a policy and performing roll-outs, the proposed
approach searches for the best model parameters so that the simulated
trajectories are as close as possible to the real ones. This is
performed through an anytime black-box Bayesian optimization, where a
belief on the optimal model is dynamically updated. Once a model with
a high probability is identified, a policy search subroutine takes over
the returned model and computes a policy to perform the target
task. The subroutine could be a control method, such as a
linear–quadratic regulator (LQR), or an RL algorithm that runs on the
physics engine with the identified model. The obtained policy is then
executed on the real robot and the newly observed trajectories are fed
to the model identification module, to repeat the same process.

The question that arises is how accurate the identified model
should be to find a successful policy. Instead of spending time searching for
the most accurate model, the optimization can stop whenever a model
that is sufficiently accurate to find a good policy is
identified. Answering this question exactly, however, is difficult,
because that would require knowing in advance the optimal policy for
the real system.


The proposed solution is motivated by a key quality desired in many
robot RL algorithms. To ensure safety, most robot RL algorithms
constrain the changes in the policy between two iterations to be
minimal and gradual. For instance, both Relative-Entropy Policy Search
(REPS)~\cite{Peters+MA:2010} and Trust Region Policy Optimization
(TRPO)~\cite{icml2015_schulman15} algorithms guarantee that the KL
divergence between an updated policy and the one in the previous
iteration is bounded. Therefore, one can in practice use the previous
best policy as a proxy to verify if there is a consensus among the
most likely models on the value of the best policy in the next iteration. This is
justified by the fact that the updated policy is not too different
from the previous one. Thus, model identification is stopped whenever
the most likely models predict almost the same value for the
previously computed policy, i.e., if all the high-probability models
predict a similar value for the previous policy, then any of these
models could be used for searching for the next policy.

Empirical evaluation performed in simulation and on a real robot show
the benefits of the framework. The initial set of experiments are
performed in the OpenAI Gym~\cite{brockman2016openai} with the MuJoCo
simulator\footnote{MuJoCo: \url{www.mujoco.org}}. They demonstrate
that the proposed model identification approach is more time-efficient
than alternatives, and it improves the data-efficiency of policy
search algorithms, such as TRPO. The second part is performed on a
real-robot manipulation task. It shows that learning using a simulator
with the identified model can be more data-efficient than other
model-based methods, such as PILCO, and model-free ones, such as
PoWER.

\section{Related Work}


This work relates to \emph{model-based RL}, where a dynamical system
is physically simulated.  Model-based learning involves explicitly
learning the unknown system dynamics and searching for an optimal
policy. Variants have been applied in
robotics~\cite{Dogar_2012_7076,LunchMason1996,isbell:physics:2014,ZhouPBM16},
where using inaccurate models can still allow a policy to steer an RC
car~\cite{abbeel2006using}. For general-purpose model-based RL, the
Probabilistic Inference for Learning COntrol (PILCO) method was able
to utilize a small amount of data to learn dynamical models and
optimal policies~\cite{Deisenroth:2011fu}. Replacing gradient-based
optimization with a parallel, black-box algorithm (Covariance Matrix
Adaptation (CMA))~\cite{hansen2006cma}, Black-DROPS can be as
data-efficient as PILCO without imposing constraints on the reward
function or the policy~\cite{chatzilygeroudis:hal-01576683}. CMA was
also used for automatically designing open-loop reference
trajectories~\cite{tan2016simulation}. Trajectory optimization was
performed in a physical simulator before real experiments, where robot
trajectories were used to optimize simulation parameters.

The proposed method also relates to \emph{learning from simulation}.
Utilizing simulation to learn a prior before learning on the robot is
a common way to reduce real-world data needs for policy search and
provides data-efficiency over PILCO~\cite{cutler2015efficient}. Policy
Improvement with REsidual Model learning
(PI-REM)~\cite{saveriano2017data} has shown improvement over PILCO by
utilizing a simulator and only modeling the residual dynamics between
the simulator and reality. Similarly, using data from simulation as
the mean function of a Gaussian Process (GP) and combining with
Iterative Linear Quadratic Gaussian Control (ILQG), GP-ILQG is able to
improve incorrect models and generate robust
controllers~\cite{lee2017gp}.

In addition, the proposed method utilizes \emph{Bayesian optimization
  (BO)} to efficiently identify the model.  For direct policy search,
a prior can also be selected from a set of candidates using
BO~\cite{pautrat2017bayesian}. BO can also be used to optimize the
policy while balancing the trade-off between simulation and real
data~\cite{DBLP:conf/icra/MarcoBHS0ST17}, or learn a locally linear
dynamical model, while aiming to maximize control
performance~\cite{bansal2017goal}.

 


\emph{Traditional system identification} builds a dynamics model by
minimizing prediction error (e.g., using least
squares)~\cite{swevers1997optimal}. There have
been attempts to combine parametric rigid body dynamics model with
nonparametric model learning for approximating the inverse
dynamics~\cite{nguyen2010using}. In contrast to these methods, this
work uses a physics engine, and concentrates on identifying model
parameters instead of learning the models from scratch. Optimizing the
simulated model parameters to match actual robot data can also help model robot damage~\cite{bongard2006resilient}. Recent
work performed in simulation proposed model identification for
predicting physical parameters, such as mass and
friction~\cite{Yu-RSS-17}.



Different from model-based methods, \emph{model-free methods} search
for a policy that best solves the task without explicitly learning the
system's
dynamics~\cite{Sutton:1998:IRL:551283,Bagnell_2013_7451}. They have
been shown effective in robotic tasks~\cite{Peters+MA:2010}.  Policy
learning by Weighting Exploration with the Returns
(PoWER)~\cite{kober2009policy} is a model-free policy search approach
widely used for learning motor skills. The TRPO
algorithm~\cite{icml2015_schulman15} has been demonstrated in some
MuJoCo simulator environments. \emph{End-to-end learning} is an
increasingly popular framework
~\cite{agrawal2016learning,WuYLFT15,byravan2017se3,finn2016deep},
which involves successful demonstrations of physical interaction and
learning a direct mapping of the sensing input to controls. These
approaches usually require many physical experiments to effectively
learn.  Overall, model-free methods tend to require a lot of training
data and can jeopardize robot safety.

A \emph{contribution} of this work is to link model identification
with policy search. It does not search for the most accurate model
within a time budget but for an accurate enough model, given an easily
computable criterion, i.e., predicting a policy's value function that
is not too different from the searched policy. This idea can be used
with many policy search algorithms that allow for smooth changes
during learning.

\section{Proposed Approach}
This section provides first a system overview and then the main
algorithm, focusing on model identification.
\subsection{System Overview and Notations}

The focus is on identifying physical properties of robots or objects,
e.g. mass and friction, using a simulator. These physical properties
are concatenated in a single vector and represented as a
$D$-dimensional vector $\theta \in \Theta$, where $\Theta$ is the
space of possible physical parameters. $\Theta$ is discretized with a
regular grid resolution. The proposed approach returns a distribution
$P$ on a discretized $\Theta$ instead of a single point $\theta \in
\Theta$ given the challenge of perfect model identification. In other
terms, there may be multiple models that can explain an observed
movement of the robot with similar accuracy. The approach is to
preserve all possible explanations and their probabilities.

The online model identification takes as input a prior distribution
$P_t$, for time-step $t\geq 0$, on the discretized parameter space
$\Theta$.  $P_t$ is calculated based on the initial distribution $P_0$
and a sequence of observations $(x_0,\mu_0, x_1,\mu_1, \dots, x_{t-1},
\mu_{t-1}, x_{t})$. For instance, $x_t$ can be the state of the robot
at time $t$ and $\mu_{t}$ is a vector describing a vector of torques
applied to the robot at time $t$. Applying $\mu_t$ results in changing
the robot's state from $x_{t}$ to $x_{t+1}$. The algorithm returns a
distribution $P_{t+1}$ on the models $\Theta$.


The robot's task is specified by a reward function $R$ that maps
state-actions $(x,\mu)$ into real numbers. A policy $\pi$ returns an
action $\mu=\pi(x)$ for state $x$. The value $V^{\pi}(\theta)$ of
policy $\pi$ given model $\theta$ is defined as $V^{\pi}(\theta) =
\sum_{t=0}^{H} R(x_t,\mu_t)$, where $H$ is a fixed horizon, $x_0$ is a
given starting state, and $x_{t+1} = f(x_{t}, \mu_t, \theta)$ is the
predicted state at time $t+1$ after simulating action $\mu_t$ in state
$x_t$ given parameters $\theta$. For simplicity, the focus is on
systems with deterministic dynamics.

\subsection{Main Algorithm}

Given a reward function $R$ and a simulator with model parameters
$\theta$, there are many ways to search for a policy $\pi$ that
maximizes value $V^{\pi}(\theta)$. For example, one can use
Differential Dynamic Programming (DDP), Monte Carlo (MC) methods, or
if a good policy cannot be found with alternatives, execute a
model-free RL algorithm on the simulator.  The choice of a particular
policy search method is open and depends on the task.  The main loop
of the system is presented in Algorithm~\ref{main_algo}. This
meta-algorithm consists in repeating three main steps: (1) data
collection using the real robot, (2) model identification using a
simulator, and (3) policy search in simulation using the best-identified model.

\begin{figure}
  \removelatexerror
\centering
\begin{minipage}[t]{0.5\textwidth}
  \begin{algorithm}[H]
$t \leftarrow 0$\;
Initialize distribution $P$ over $\Theta$ to a uniform distribution\;
Initialize policy $\pi$\;
\Repeat{ Timeout}{
Execute policy $\pi$ for $H$ iterations on the real robot, 
and collect new state-action-state data $\{(x_{i},\mu_{i}, x_{i+1})\}$ for $i=t, \dots,t+H-1$\;
$t \leftarrow t+H$\;
Run Algorithm~\ref{greedyES_Alg} with collected state-action-state data and reference policy $\pi$
for updating distribution $P$\;
Initialize a policy search algorithm (e.g. TRPO) with $\pi$ and run the algorithm in the simulator with the model $\arg\max_{\theta\in \Theta} P(\theta)$
to find an improved policy $\pi'$;\
$\pi \leftarrow \pi'$\;
}
\caption{Main Loop}
\label{main_algo}
\end{algorithm}
\end{minipage}
\end{figure}

\subsection{Value-Guided Model Identification}
\label{model_id}
The process, explained in Algorithm~\ref{greedyES_Alg}, consists of simulating the effects of actions $\mu_{i}$ on the robot in states $x_{i}$ under various values of parameters $\theta$ and observing the resulting states $\hat{x}_{i+1}$, for $i=0, \dots,t$. The goal is to identify the model parameters that make the outcomes of the simulation as close as possible to the real observed outcomes.  In other terms, the following black-box optimization problem is solved: 
\begin{eqnarray}
\theta^* = \arg \min_{\theta \in \Theta} E(\theta)  \stackrel{def}{=}  \sum_{i=0}^{t}\|x_{i+1} - f(x_{i}, \mu_i, \theta) \|_2,
\label{gpError}
\end{eqnarray}
wherein $x_{i}$ and $x_{i+1}$ are the observed states of the robot at times $i$ and $i+1$,  $\mu_i$ is the action that moved the robot from  $x_{i}$ to $x_{i+1}$, and $f(x_{i}, \mu_i, \theta) =  \hat{x}_{i+1}$, the predicted state at time $i+1$ after simulating action $\mu_i$ in state $x_i$ using $\theta$.

\begin{figure}[t!]
  \removelatexerror
\centering
\begin{minipage}[t]{0.5\textwidth}
  \begin{algorithm}[H]


\KwIn{state-action-state data $\{(x_{i},\mu_{i}, x_{i+1})\}$ for $i=0, \dots,t$\newline
 a discretized space of possible values of physical properties  $\Theta$, \newline
a reference policy $\pi$,\newline
minimum and maximum number of evaluated models $k_{min},k_{max}$, \newline
model confidence threshold $\eta$, \newline 
value error threshold $\epsilon$ \;}
\KwOut{probability distribution $P$ over $\Theta$\;}
Sample $\theta_0 \sim \textrm{Uniform}(\Theta)$; $L \leftarrow \emptyset$; $k\leftarrow 0$; $stop \leftarrow false$\;
\Repeat{ stop = true}{
    \tcp{\small Calculating the accuracy of model $\theta_k$}
$l_k \leftarrow 0$\;
\For{$i = 0$ \KwTo $t$}{
Simulate $\{(x_{i},\mu_{i})\}$ using a physics engine with physical parameters $\theta_k$ and get the predicted next state $\hat{x}_{i+1} = f(x_{i}, \mu_i, \theta_k)$ \;
$l_k \leftarrow l_k + \|\hat{x}_{i+1} - x_{i+1}\|_2$\;
}
$L \leftarrow L \cup \{(\theta_k , l_k)\}$\;
Calculate $GP(m,K)$ on error function $E$, where $E(\theta) = l$, using data $(\theta , l) \in L$\;
   \tcp{\small Monte Carlo sampling}
Sample $E_1,E_2, \dots, E_n \sim GP (m,K)$ in $\Theta$\;
\ForEach{$\theta \in \Theta$}{
\begin{eqnarray}P (\theta) \approx \frac{1}{n}\sum_{j=0}^{n} \mathbf{1}_{ \theta = \arg \min_{\theta' \in \Theta} E_j(\theta') }\label{MC_sampling}\end{eqnarray}
}
   \tcp{\small Selecting the next model to evaluate}
$\theta_{k+1} = \arg \min_{\theta\in \Theta} P(\theta) \log \big(P (\theta)\big)$ \;
 $k\leftarrow k+1$\;
   \small \tcp{ Checking the stopping condition}
       
     \If{$k \geq k_{min} $}{
         $\theta^*\leftarrow\arg\max_{\theta\in \Theta} P(\theta)$\;        
         Calculate the values $V^{\pi}(\theta)$ with all models $\theta$ that have a probability $P(\theta)\geq \eta$ by using the physics engine for simulating 
         trajectories with models $\theta$  \;
        \If{$\sum_{\theta, \forall P(\theta)\geq \eta } P(\theta) |V^{\pi}(\theta) - V^{\pi}(\theta^*)| \leq \epsilon$}{ $stop \leftarrow true $\;}
     }
     \If{$k = k_{max} $}{ $stop \leftarrow true $\;}

}
\caption{Value-Guided Model Identification (VGMI)}
\label{greedyES_Alg}
\end{algorithm}
\end{minipage}
\end{figure}

High-fidelity simulations are computationally expensive. Thus, when
searching for the optimal parameters in Eq.~\ref{gpError}, it is
important to minimize the number of evaluations of function $E$. This
is achieved through {\it Entropy Search}~\cite{HennigSchuler2012},
which explicitly maintains a belief on the optimal parameters. This is
unlike other Bayesian optimization methods, such as {\it Expected
  Improvement}, which only maintains a belief on the objective
function. The following description explains how this technique is
adapted so as to provide an automated stopping criterion.

The error function $E$ does not have an analytical form but is learned
from a sequence of simulations with a small number of parameters
$\theta_k\in \Theta$. To choose these parameters efficiently in a way
that quickly leads to accurate parameter estimation, a belief about
the actual error function is maintained. This is a probability measure
over the space of all functions $E : \mathbb{R}^D \rightarrow
\mathbb{R}$, represented by a Gaussian Process
(GP)~\cite{RasmussenGPM} with mean vector $m$ and covariance matrix
$K$. The mean $m$ and covariance $K$ of the GP are learned from data:
$\{\big(\theta_0, E(\theta_0)\big), \dots, \big(\theta_k,
E(\theta_k)\big)\}$, where $\theta_k$ is a vector of physical robot
properties, and $E(\theta_k)$ is the accumulated distance between
observed states and simulated states given $\theta_k$.

The probability distribution $P$ on the identity of the best physical model $\theta^*$, returned by the algorithm, is computed from the learned GP as
\begin{eqnarray}
\begin{aligned}
P(\theta) &\stackrel {def}{=} P\big(\theta = \arg\min_{\theta' \in \Theta} E(\theta')\big)\\
&=  \int_{E: \mathbb{R}^D \rightarrow \mathbb{R}}  p_{m,K}(E) \Pi_{\theta' \in \Theta-\{\theta\} } H \big(E(\theta') - E(\theta)\big)  \mathrm{d}E
\end{aligned}
\label{minDistribution}
\end{eqnarray}

where $H$ is the Heaviside step function, i.e., $H \big(E(\theta') - E(\theta)\big) = 1$ if $E(\theta') \geq E(\theta)$ and  $H \big(E(\theta') - E(\theta)\big) = 0$ otherwise, and  $p_{m,K}(E)$ is the probability of $E$ according to the learned GP mean $m$ and covariance $K$. Intuitively, $P(\theta)$  is the expected number of times that $\theta$ happens to be the minimizer of $E$ when $E$ is a function distributed according to the GP. 

The distribution $P$ from Eq.~\ref{minDistribution} does not have
a closed-form expression. Therefore, {\it Monte Carlo} (MC) sampling
is employed for estimating $P$.  The process samples vectors
$[E(\theta')]_{\theta' \in\Theta}$ containing values that $E$ could
take, according to the learned GP, in the discretized
space $\Theta$. Then $P(\theta)$ is estimated by counting the ratio of
sampled vectors of the values of simulation error $E$ where $\theta$
happens to make the lowest error, as indicated in
Eq.~\ref{MC_sampling} in Alg.~\ref{greedyES_Alg}.

Finally, the computed distribution $P$ is used to select the next vector $\theta_{k+1}$ to use as a physical model in the simulator. This process is repeated until the entropy of $P$ drops below a certain threshold, or until the algorithm runs out of the allocated time budget.  The entropy of $P$ is given as $\sum_{\theta\in \Theta} -P_{min}(\theta) \log \big(P_{min}(\theta)\big)$. When the entropy of $P$ is close to zero, the mass of distribution $P$ is concentrated around a single vector $\theta$, corresponding to the physical model that best explains the observations. Hence, next $\theta_{k+1}$ should be selected so that the entropy of $P$ would decrease after adding the data point $\big(\theta_{k+1}, E(\theta_{k+1})\big)\}$ to train the GP and re-estimate $P$ using the new mean $m$ and covariance $K$ in Eq.~\ref{minDistribution}. 

Entropy Search methods follow this reasoning and use MC again to
sample, for each potential choice of $\theta_{k+1}$, a number of
values that $E(\theta_{k+1})$ could take according to the GP in order
to estimate the expected change in the entropy of $P$ and choose the
parameter vector $\theta_{k+1}$ that is expected to decrease the
entropy of $P$ the most. The existence of a secondary nested process
of MC sampling makes this method impractical for online
optimization. Instead, this work presents a simple heuristic for
choosing the next $\theta_{k+1}$. In this method, called {\it Greedy
  Entropy Search}, the next $\theta_{k+1}$ is chosen as the point that
contributes the most to the entropy of $P$,

\begin{eqnarray}
\theta_{k+1} = \arg \max_{\theta\in \Theta} - P(\theta) \log \big(P (\theta)\big).
\end{eqnarray}

This selection criterion is greedy because it does not anticipate how
the output of the simulation using $\theta_{k+1}$ would affect the
entropy of $P$. Nevertheless, this criterion selects the point that is
causing the entropy of $P$ to be high. That is a point $\theta_{k+1}$
with a good chance $P(\theta_{k+1})$ of being the real model, but also
with a high uncertainty $P(\theta_{k+1}) \log
\big(\frac{1}{P(\theta_{k+1})}\big)$. Initial experiments suggested
that this heuristic version of Entropy Search is more practical than
the original Entropy Search method because of the computationally
expensive nested MC sampling loops used in the original method. The
actual sampled values are not restricted to lie on the discretized
grid. Once a grid value is selected as the candidate with the highest
entropy, a local optimization process using L-BFGS takes place to
further optimize the parameter value.

\begin{figure*}[!ht]
        \centering
       \includegraphics[width=0.3\textwidth]{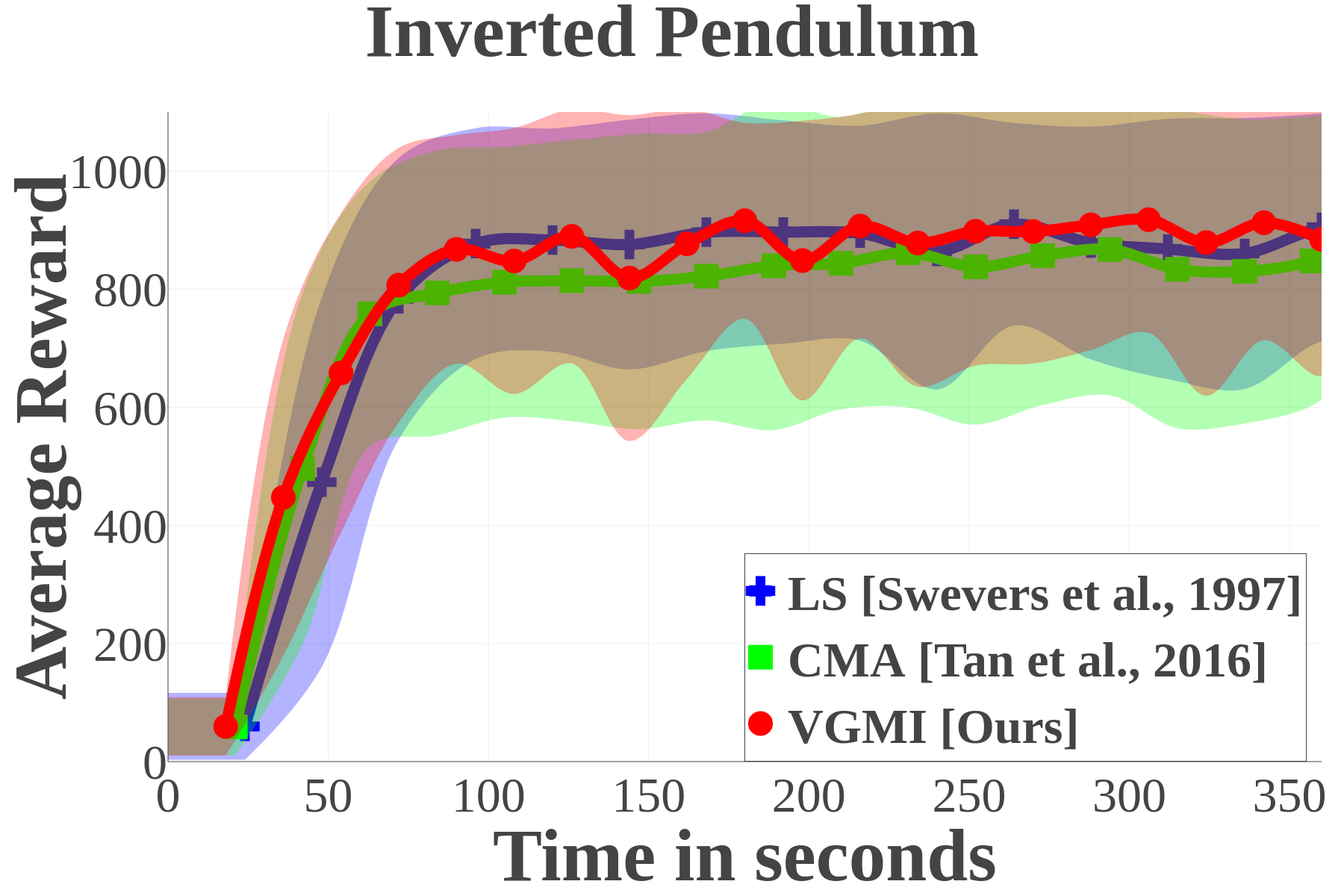} 
       \includegraphics[width=0.3\textwidth]{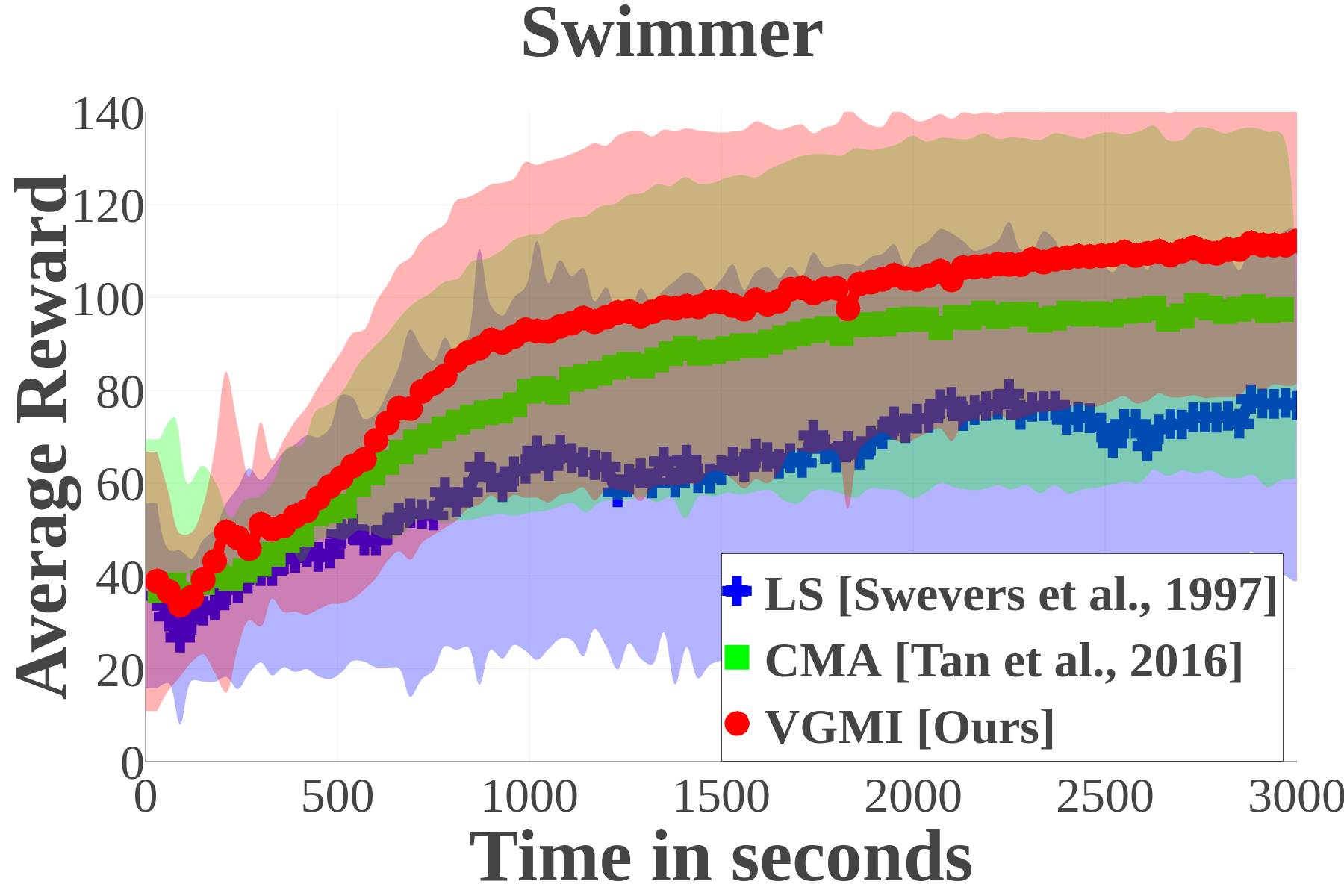}   
       \includegraphics[width=0.3\textwidth]{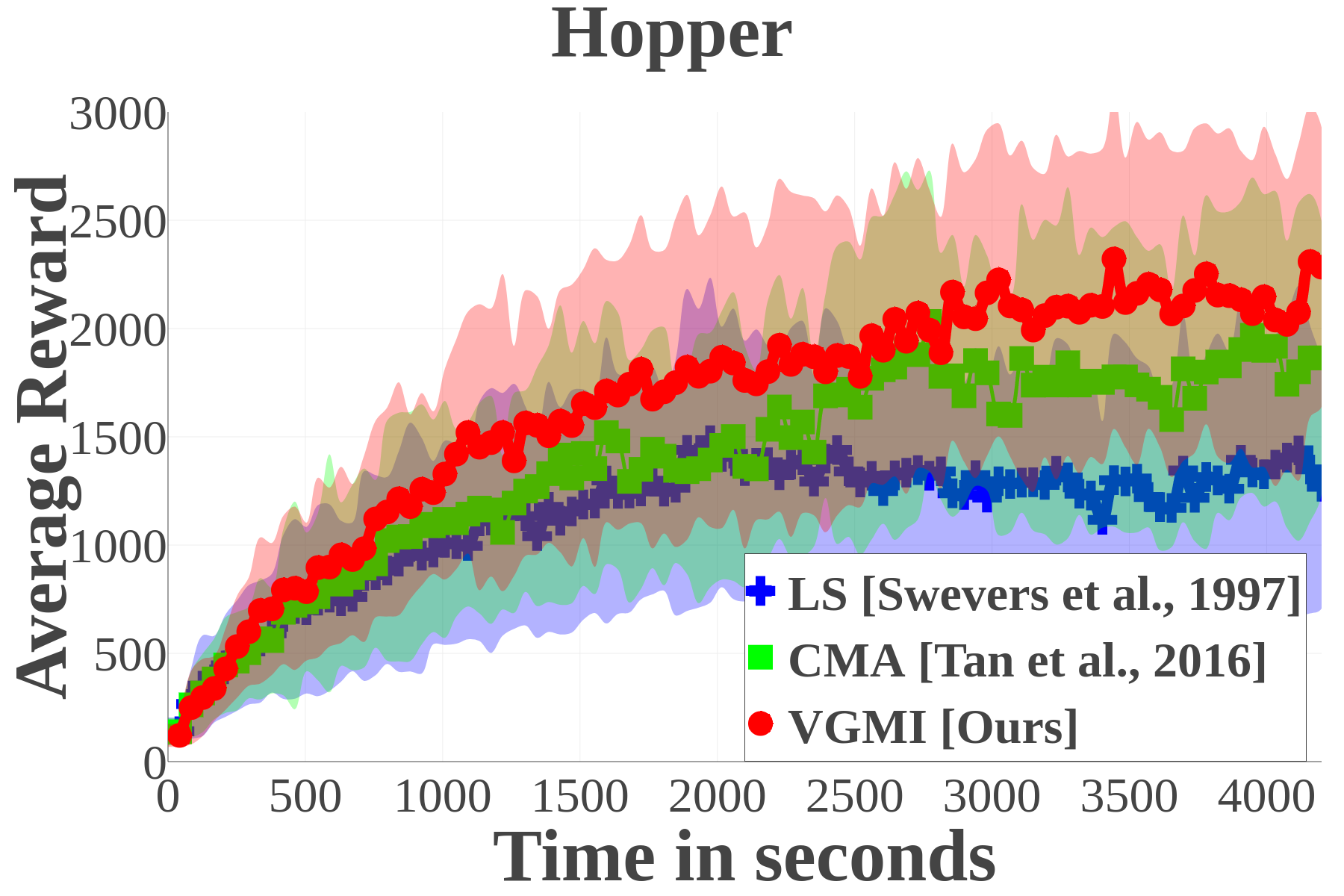}   
       \includegraphics[width=0.3\textwidth]{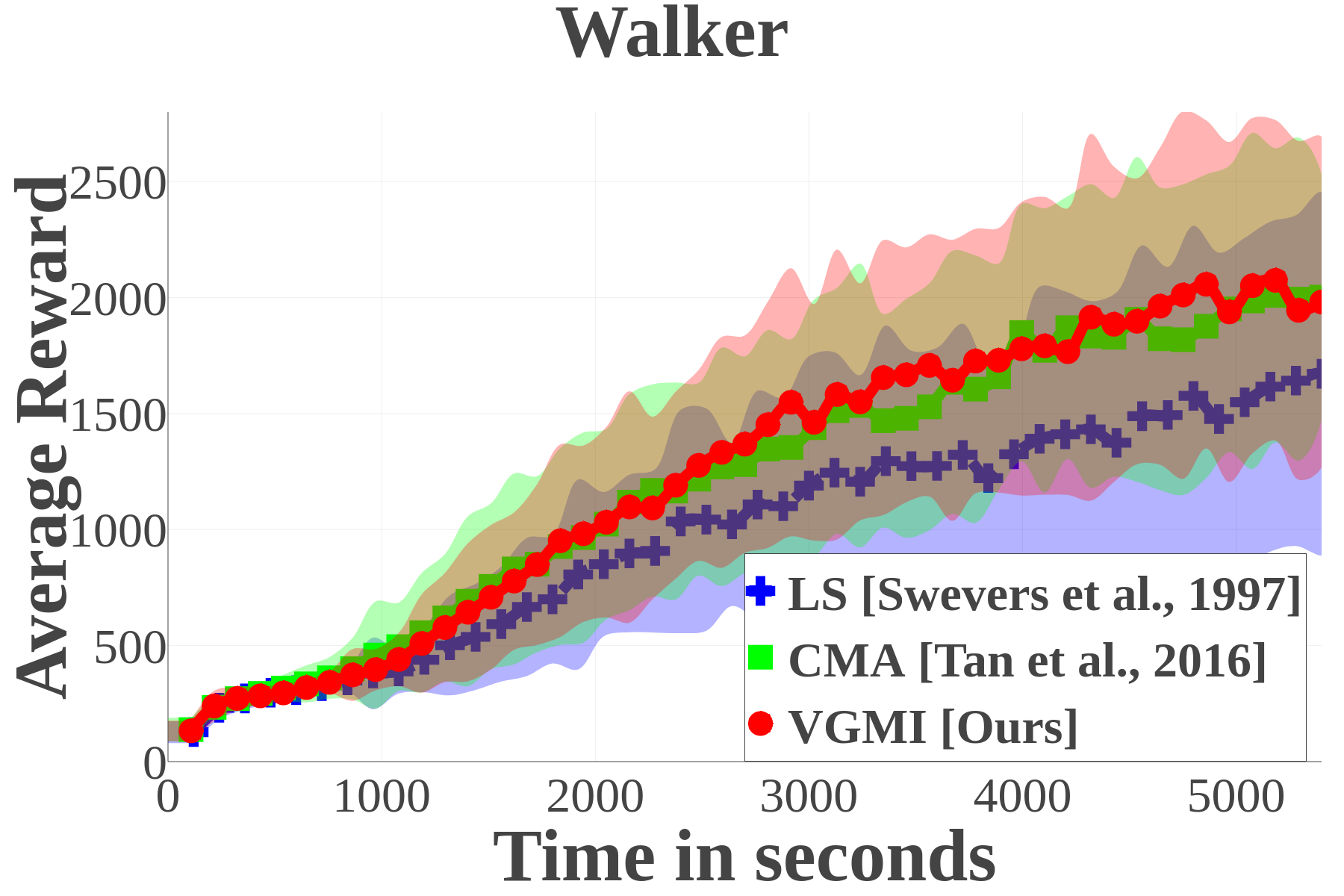}   
       \includegraphics[width=0.3\textwidth]{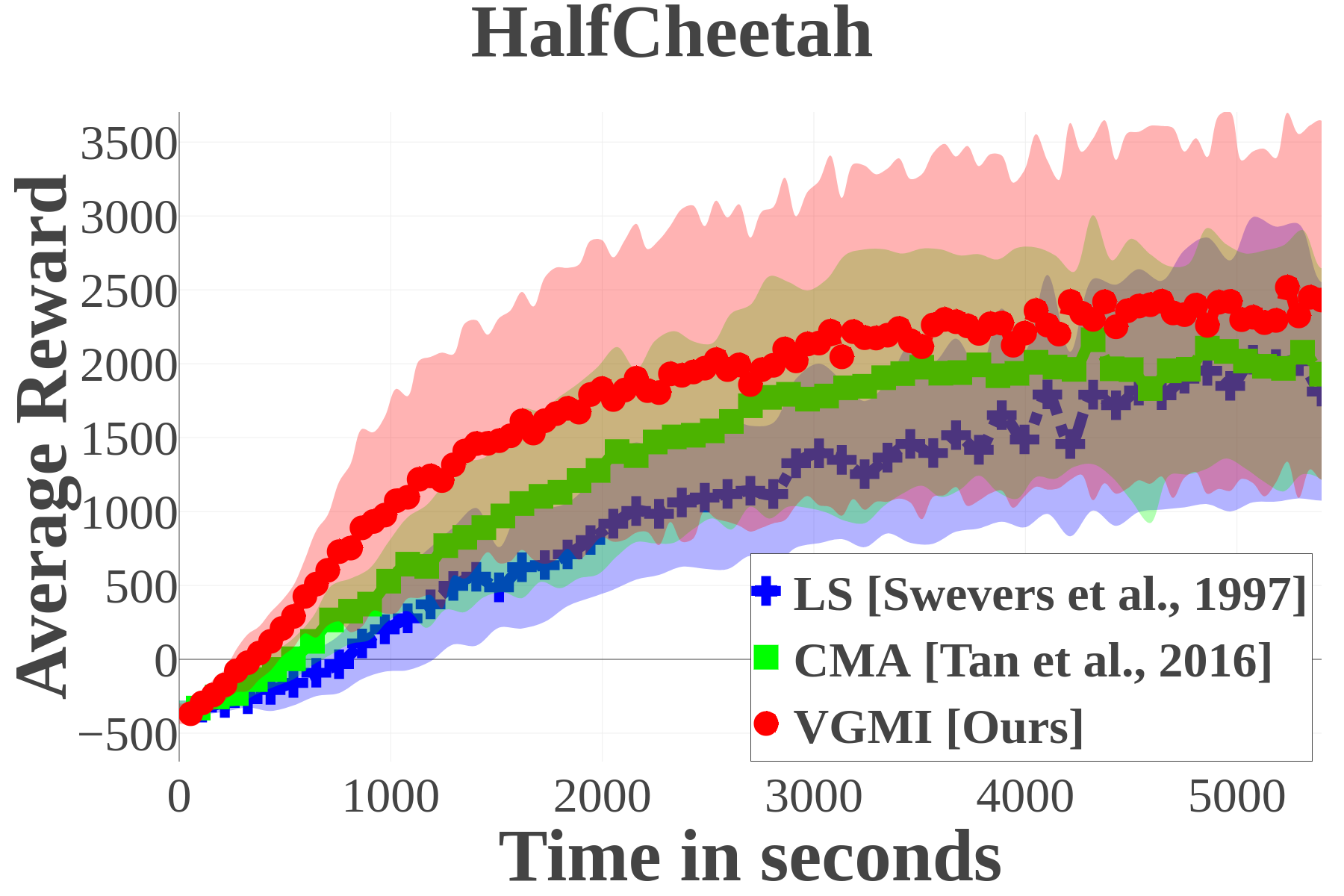}
\caption{\small Average reward as a function of
  total time, including both model identification and policy search. (Best viewed in color.)}
\label{fig:SimulationTime}
\end{figure*}

\begin{figure*}[!ht]
        \centering
       \includegraphics[width=0.26\textwidth]{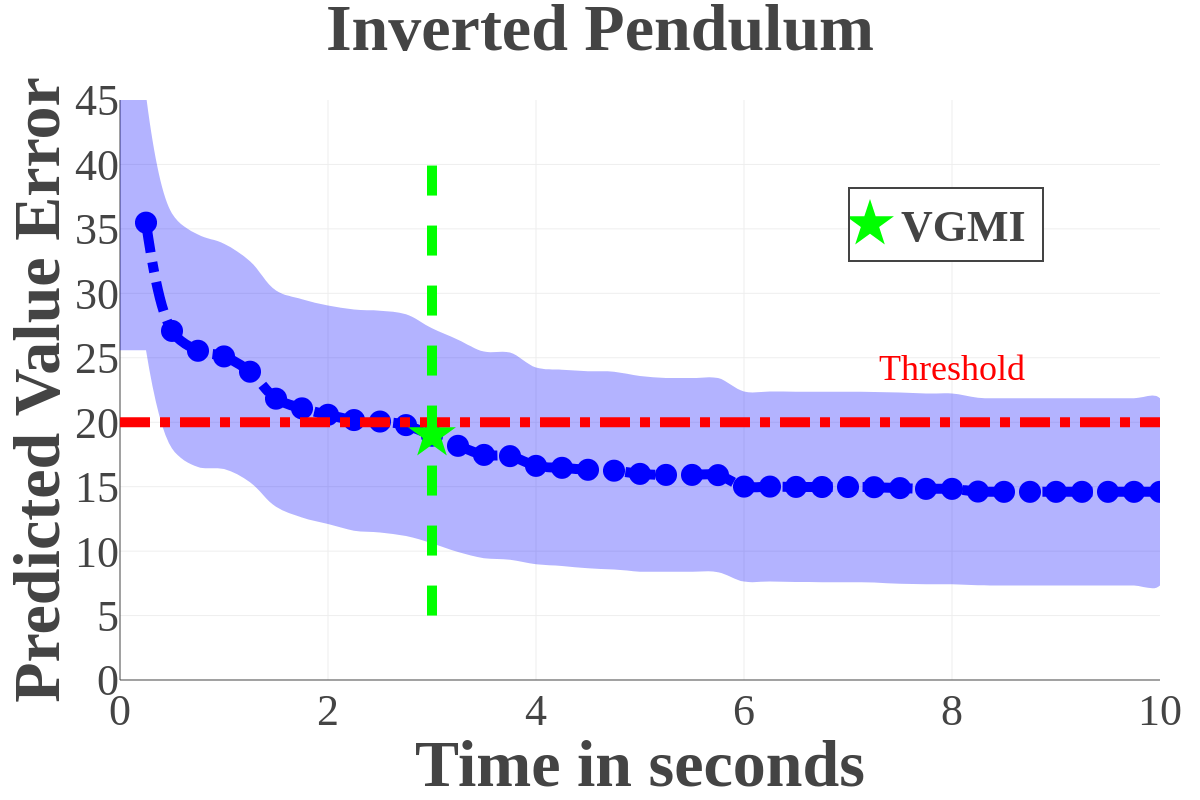} 
       \includegraphics[width=0.26\textwidth]{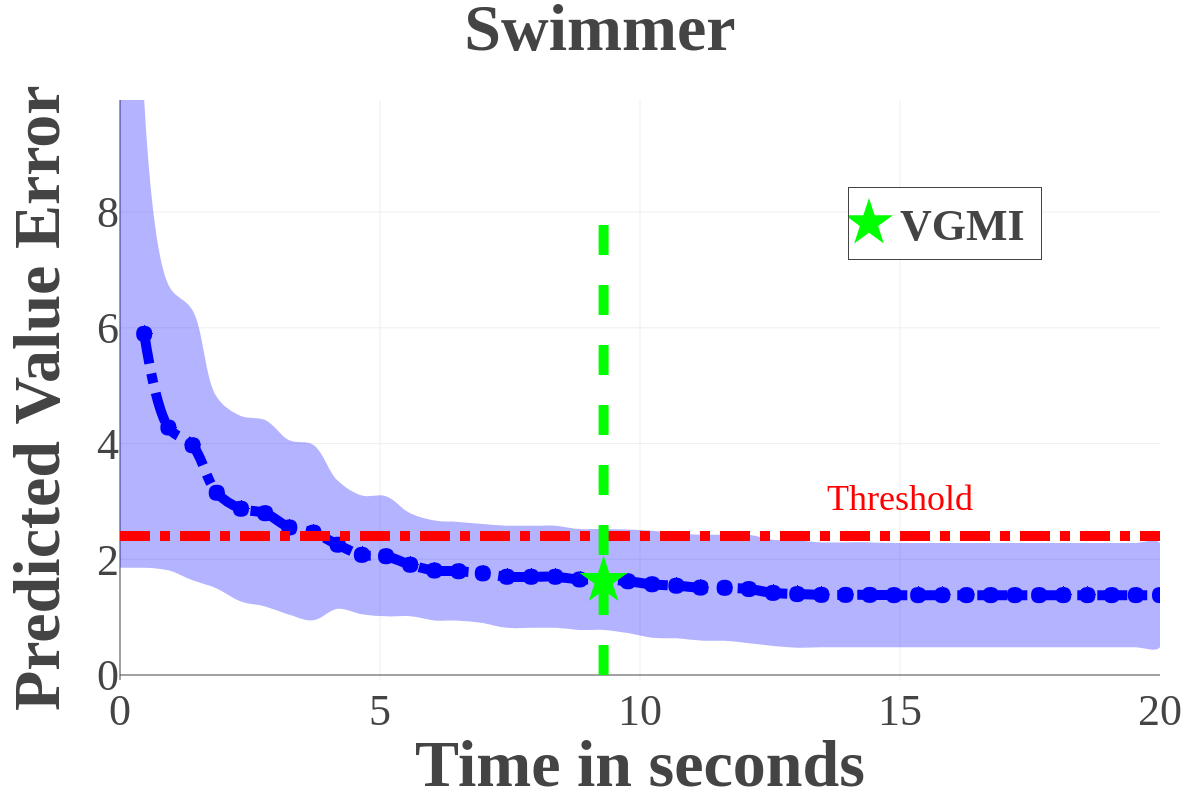}   
       \includegraphics[width=0.26\textwidth]{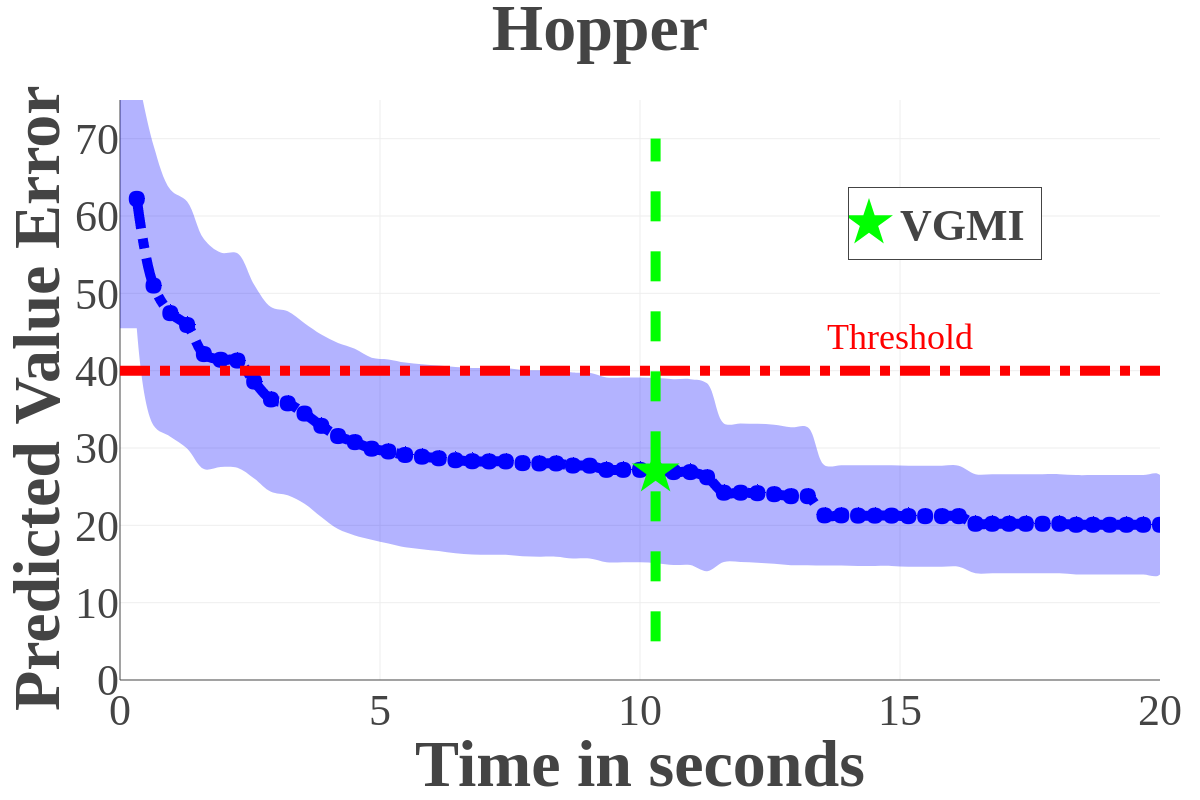}   
       \includegraphics[width=0.26\textwidth]{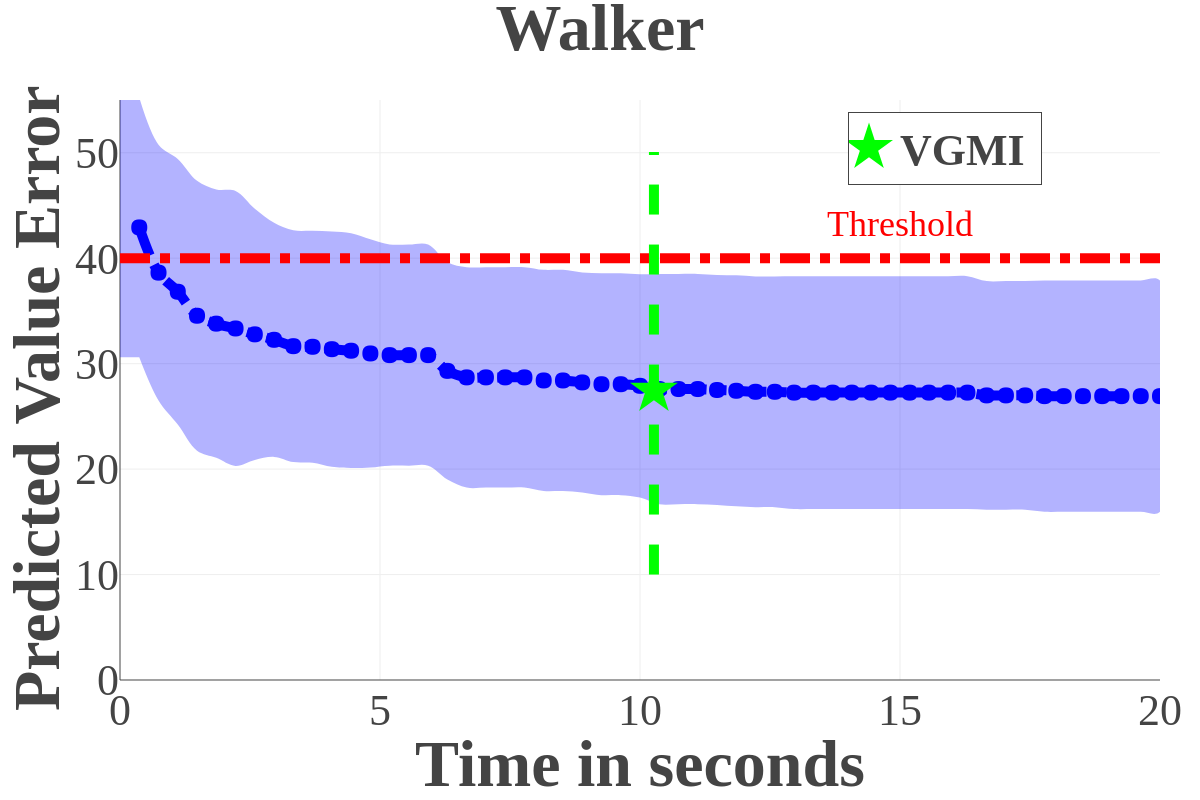}
       \includegraphics[width=0.26\textwidth]{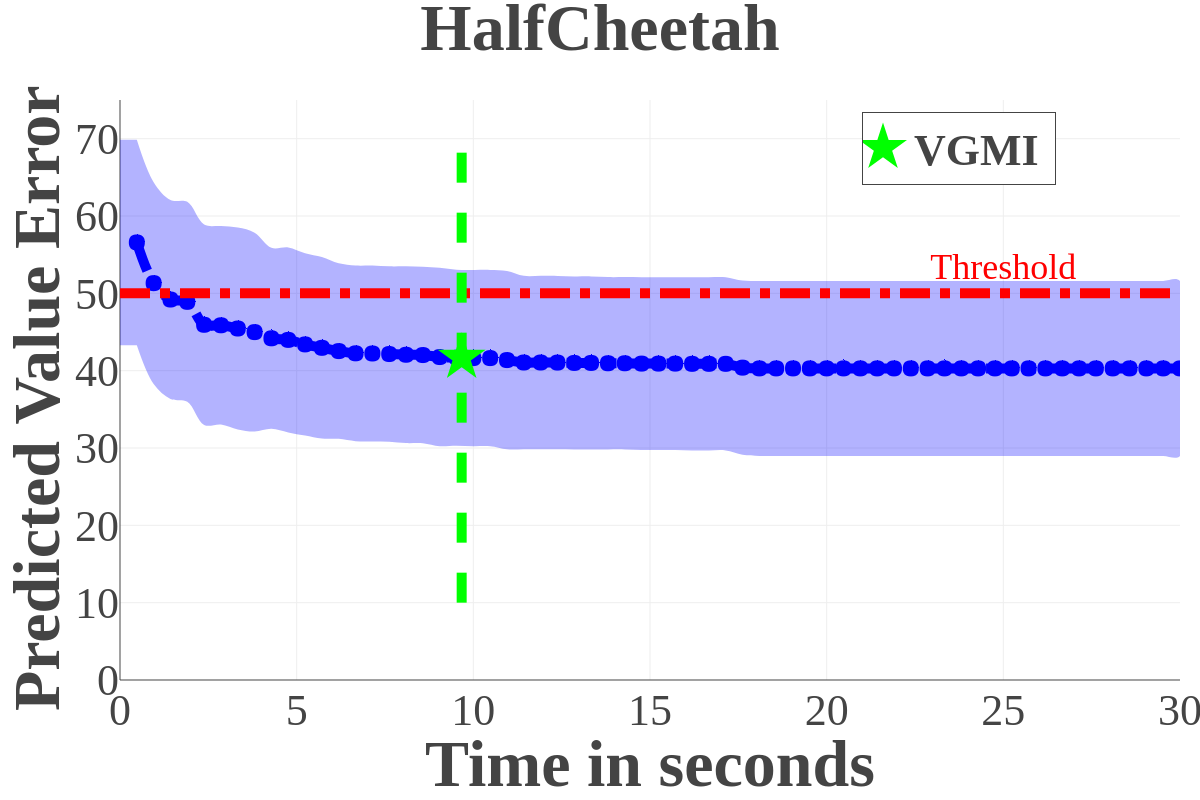}
       \caption{\small  Error ($|V^{\pi'}(\theta) - V^{\pi'}(\theta^*)|$) in predicting the value of the next policy ${\pi'}$ of TRPO as a function of the time spent on model identification. This error is unknown to the VGMI algorithm, because it is computed by using the ground-truth model $\theta^*$ and comparing its prediction against the prediction of the most probable model $\theta$ so far returned by the algorithm.
       This error is unknown to VGMI also because the next policy ${\pi'}$ to be returned by TRPO is unknown, and VGMI uses the previous policy $\pi$ as a proxy for it in its stopping condition.
       We computed this error {\it a posteriori} to assess if VGMI does not stop prematurely, or too late after finding a good enough model. 
       When VGMI decides to stop (green line), the actual error is indeed always below the theoretical threshold (red line) used in its stopping condition.}
       \label{fig:VGMI_threshold}
\end{figure*}

The stopping condition of Alg.~\ref{greedyES_Alg} depends on the predicted value of a reference policy $\pi$.
The reference policy is one that will be used in the main algorithm (Alg.~\ref{main_algo}) as a starting point in the policy search with the identified model. 
That is also the policy executed in the previous round of the main algorithm. Many policy search algorithms (such as REPS and TRPO) guarantee that
the KL divergence between consecutive policies $\pi$ and $\pi'$ is minimal. Therefore, if the difference $|V^{\pi}(\theta) - V^{\pi}(\theta^*)|$ for two given models $\theta$ and $\theta^*$
is smaller than a threshold $\epsilon$, then the difference $|V^{\pi'}(\theta) - V^{\pi'}(\theta^*)|$ should also be smaller than a threshold that is a function of  $\epsilon$ and $KL(\pi\|\pi')$. 
A full proof of this conjecture is the subject of an upcoming work. In practice, this means that if $\theta$ and $\theta^*$ are two models with high probabilities, and $|V^{\pi}(\theta) - V^{\pi}(\theta^*)|\leq \epsilon$ then there is no point in continuing the optimization to find out which one of the two models is actually the most accurate because both models will result in similar policies. The same argument could be used when there are more than two models with high probabilities. 

\section{Experimental Results}
\label{sec:result}

VGMI is evaluated both in simulation and on a real robot.

\subsection{Experiments on RL Benchmarks in Simulation}

\noindent {\bf Setup}: The simulation experiments are performed in
OpenAI Gym with the MuJoCo
simulator. The space of unknown physical models $\theta$
is:

\noindent {\bf Inverted Pendulum (IP)}: A pendulum is connected
to a cart, which moves linearly. The dimensionality of $\Theta$ is two, one for the mass of the pendulum and one for the
cart.

\noindent {\bf Swimmer}: The swimmer is a 3-link planar robot. $\Theta$ has
three dimensions, one for the mass of each link.

\noindent {\bf Hopper}: The hopper is a 4-link planar mono-pod robot.
Thus, the dimensionality of $\Theta$ is four.

\noindent {\bf Walker2D}: The walker is a 7-link planar biped
robot. Thus, the dimensionality of $\Theta$ is seven.

\noindent {\bf HalfCheetah}: The halfcheetah is a 7-link planar cheetah
robot. The dimensionality of the space $\Theta$ is seven.

The simulator starts the ground truth mass, which is only used for
rollouts to generate data for model identification.  For inaccurate
simulators as priors, the mass was randomly increased or decreased by
10 to 15\%. All the policies are trained with TRPO using
rllab~\cite{duan2016benchmarking}. The policy network has 2 hidden
layers with 32 neurons each. VGMI is compared against a) Covariance
Matrix Adaptation Evolution Strategy (CMA-ES)
~\cite{tan2016simulation} and b) Least Square (LS) optimization using
L-BFGS-B ~\cite{swevers1997optimal}.  All of them optimize the
objective function of Eq.~\ref{gpError}.

\begin{figure}[!ht]
        \centering
       \includegraphics[width=0.36\textwidth]{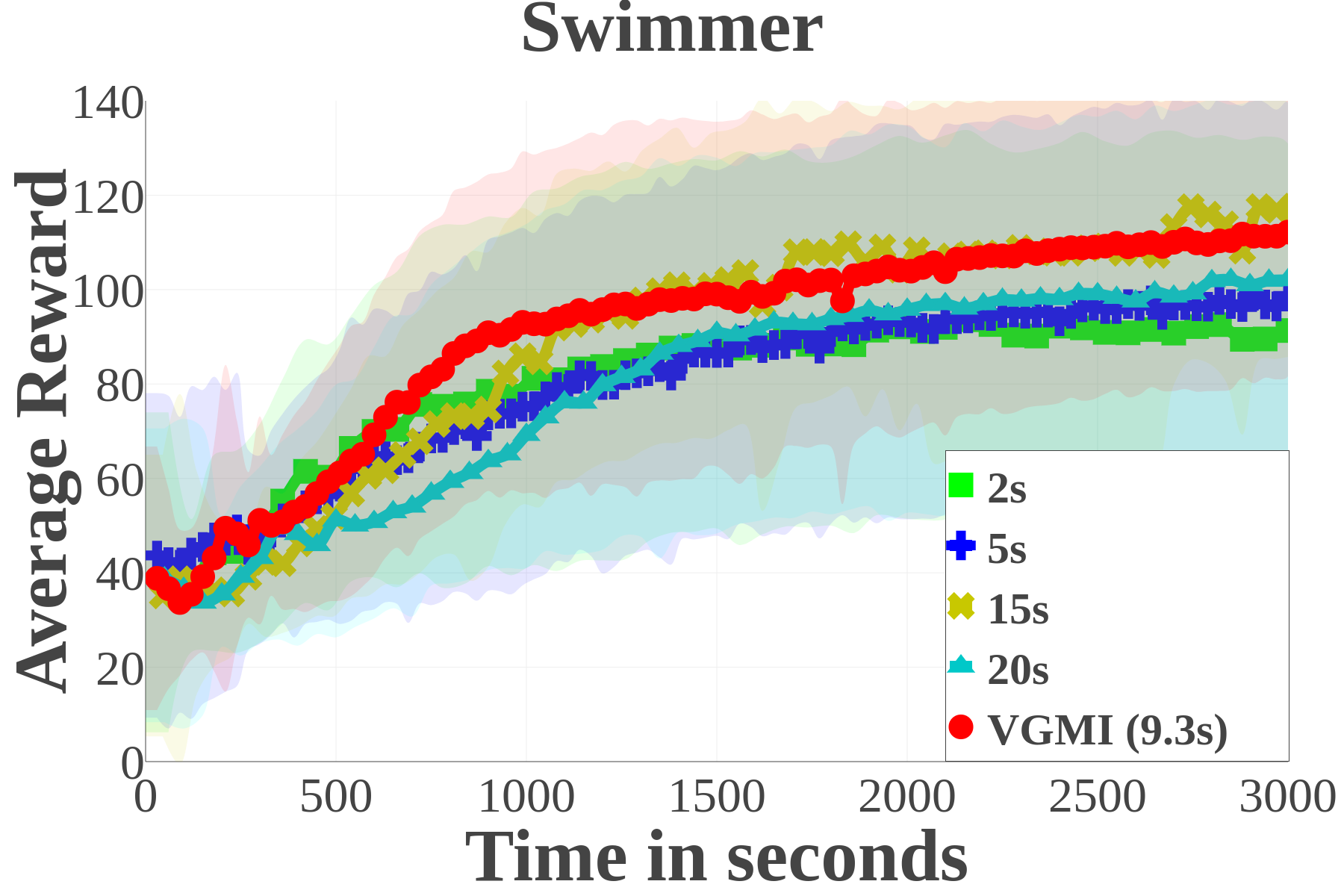} 
       \includegraphics[width=0.36\textwidth]{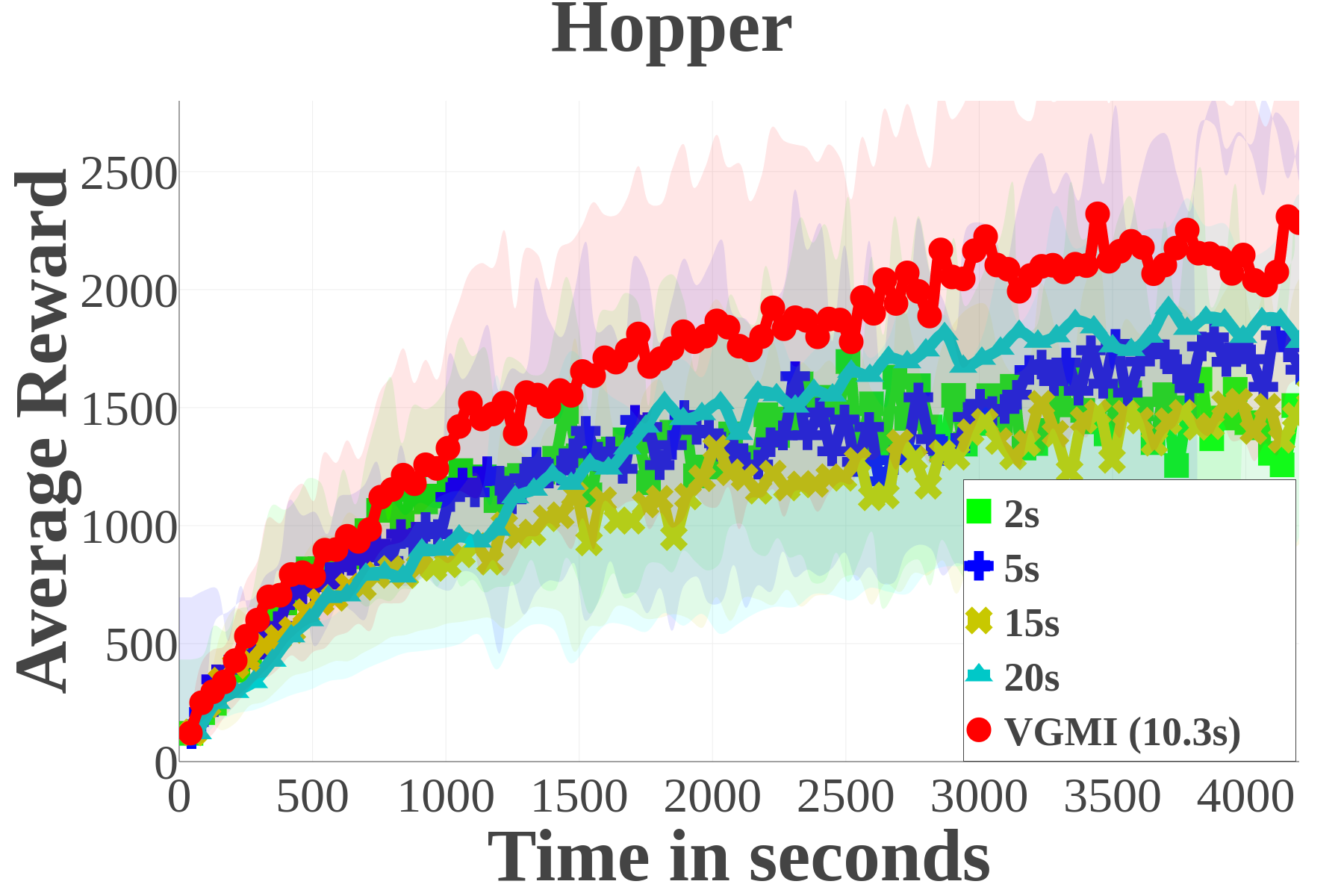}     
       \caption{\small For a fixed total time budget for both model
         identification and policy search, VGMI performs better than
         manually selected time budgets for model identification. (Best viewed in color.)}
       \label{fig:bo_time}
\end{figure}

The number of iterations for policy search varies on problem
difficulty. The main loop in Alg.~\ref{main_algo} is executed for 20
iterations for IP, 100 iterations for Swimmer, 100 iterations for
Hopper, 200 iterations for Walker2D and 400 iterations for
HalfCheetah. VGMI follows Alg.~\ref{greedyES_Alg} and is executed
every 10 iterations, i.e., $H=10$ in Alg.~\ref{main_algo}. The same
iterations are used for CMA and LS and they are also given the same
time budget for model identification.  All the results are the mean
and stand deviation of 20 independent trials.

\noindent {\bf Results:} Performance is reported for total training
time, i.e., including both model identification and policy search.
Fig. \ref{fig:SimulationTime} shows the cumulative reward per
trajectory on the ground truth system as a function of total
time. VGMI always achieves the highest reward across benchmarks and
earlier than CMA and LS. CMA performs better than LS, though relies on
a large number of physically simulated trajectories.

Fig. \ref{fig:VGMI_threshold} assesses the automated stopping of VGMI. 
The red line is the error threshold $\epsilon$ in the
predicted value. The green line is when VGMI stops
based on $\epsilon$. The same algorithm
was also executed with increasing pre-set time budgets (x-axis). The
reality gap in predicting the value of the best policy was empirically
estimated using the best-identified and the ground truth model.  This
ensures that VGMI does not stop prematurely. When it stops (green
line), the actual error (blue
curve) is indeed below the predefined threshold $\epsilon$.

\begin{figure}[!ht]
  \centering
  \vspace{-0.5in}
  \includegraphics[width=0.48\textwidth]{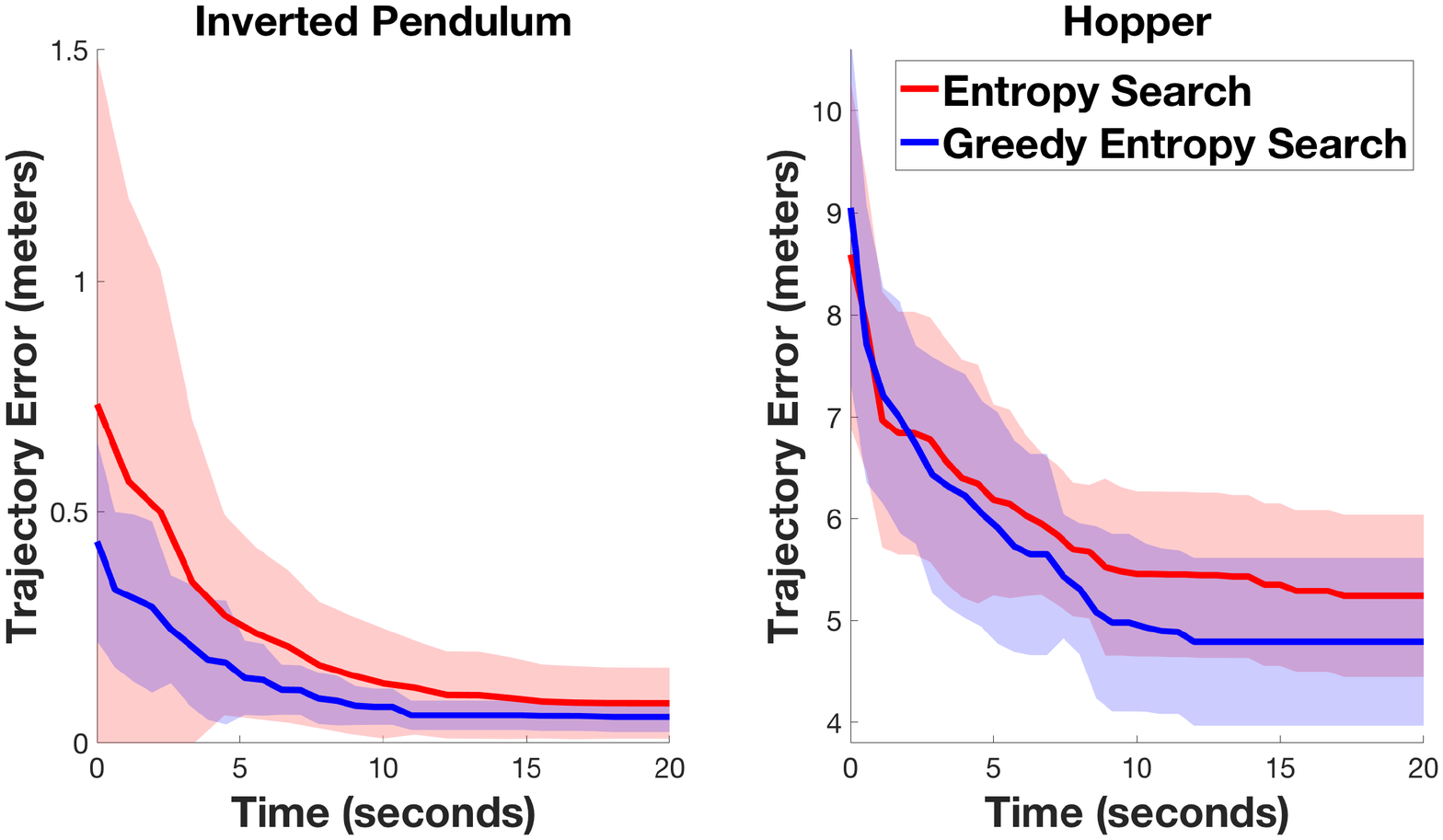}
  \vspace{-0.7in}
\caption{\small Model identification in Inverted Pendulum and Hopper environment using two variants of Entropy Search.}
\label{fig:GES}
\end{figure}

Fig.  \ref{fig:bo_time} shows that the adaptive stopping criterion of
VGMI performs better than any fixed time budget for model
identification. Thus, the proposed stopping criterion effectively
balances the trade-off between model identification and policy search.
Fig. \ref{fig:GES} demonstrates that Greedy Entropy Search (GES) in
the context of VGMI results in improved convergence relative to the
original Entropy Search.

\begin{figure*}[!ht]
\begin{center}         \includegraphics[width=\textwidth]{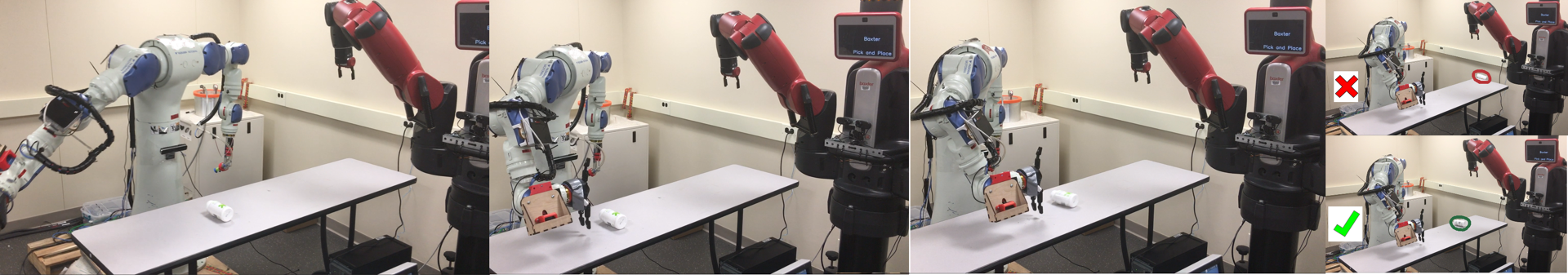}
\end{center}
\caption{\small Experiment where the {\it Motoman} pushes the object into
  Baxter's workspace after identifying the problem's physical parameters.}
\label{fig:full_push}
\end{figure*}

\begin{figure}[!ht]
\centering
\includegraphics[width=0.48\textwidth]{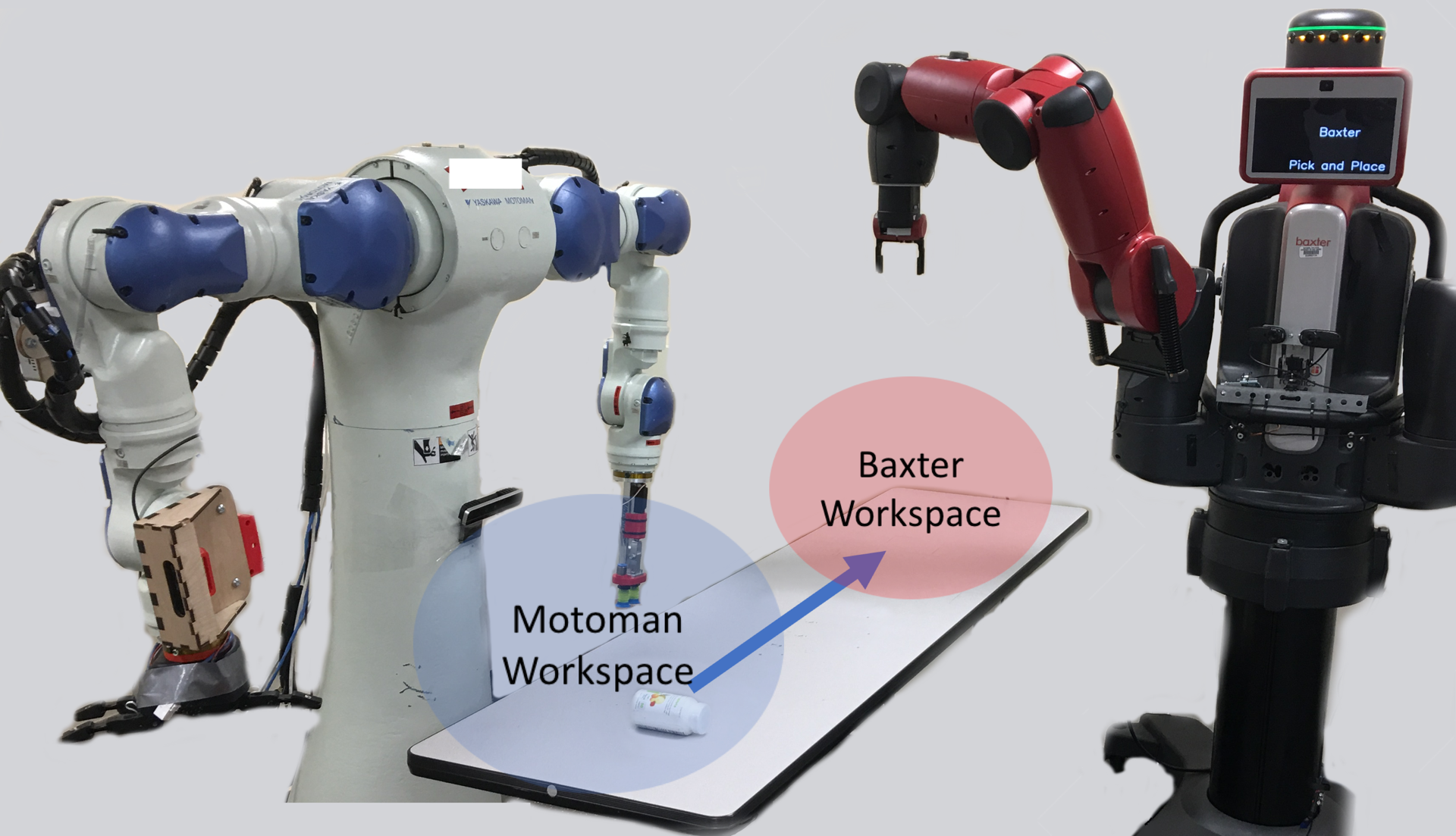}
\caption{\small Baxter needs to pick up the bottle but cannot reach it, while
  the Motoman can. Motoman gently pushes the object locally without
  losing it and identifies the object's parameters from observed
  motions and a physics engine. The object is then pushed to Baxter
  using a policy learned in simulation with the identified
  parameters.}
\label{fig:ex}
\end{figure}

\subsection{Real Robot Pushing Experiments}
\label{exp:high_spped}

\noindent {\bf Setup:} Consider Figs. \ref{fig:full_push} and
\ref{fig:ex}, where {\it Motoman} assists ({\it Baxter}) to pick up a
bottle.  The object is known but can be reached only by Motoman.  The
intersection of the robots' reachable workspace is empty and Motoman
must learn an action to push the bottle 1m away within Baxter's
reachability. If Motoman simply executes a maximum velocity push, the
object falls off the table. Similarly, if the object is rolled too
slowly, it can get stuck in the region between the two robots. Both
outcomes are undesirable as they require human intervention to reset
the scene.

The goal is to find an optimal policy with parameter $\eta$
representing the pushing velocity. The pushing direction is towards
the target and the hand pushes the object at its geometric center. No
human effort was needed to reset the scene. A pushing velocity limit
and a pushing direction were set so the object is always in Motoman's
workspace.  The approach iteratively searches for the best $\eta$ by
uniformly sampling 20 different velocities in simulation, and
identifies the object model parameters $\theta^*$ (the mass and
friction coefficient) using trajectories from rollouts by running VGMI
as in Alg.~\ref{greedyES_Alg}.  VGMI is executed after each rollout,
i.e., $H=1$ in Alg.~\ref{main_algo}. The method is compared to
PoWER~\cite{kober2009policy} and PILCO~\cite{Deisenroth:2011fu}. For
PoWER, the reward function is $r=e^{-dist}$, where $dist$ is the
distance between the object position after pushing and the desired
target. For PILCO, the state space is the 3D object position.

\begin{figure}[!ht]
        \centering
       \includegraphics[width=0.335\textwidth]{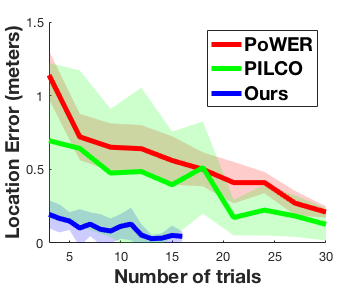}   
       \includegraphics[width=0.335\textwidth]{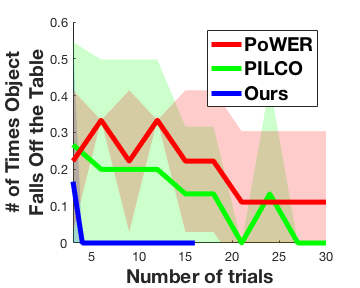} 
\caption{\small Pushing policy optimization results using a {\it Motoman}
  robot. VGMI achieves both lower final object location error and
  fewer object drops comparing to alternatives. (Best viewed in color)}
\label{fig:Rollouts}
\end{figure}

\noindent {\bf Results:} Two metrics are used for evaluating
performance: 1) The distance between the final object location after
being pushed and the desired goal location; 2) The number of times the
object falls off the table.  Fig. \ref{fig:Rollouts} shows that in the
real-world experiments the method achieves both lower final object
location error and fewer number of object drops, which is important
for robot learning that minimizes human effort. The model-free PoWER
results in higher location error and more object drops. PILCO performs
better than PoWER as it learns a dynamical model but the model is not
as accurate as the VGMI one. Since simple policy search is used for
VGMI, the performance is expected to be better in more advanced policy
search methods, such as combining PoWER with VGMI.  A video with the
experiments can be found on \url{https://goo.gl/Rv4CDa}.

\vspace{-0.1in}
\section{Conclusion}
\label{sec:conclusion}

This paper presents a practical approach that integrates physics
engines and Bayesian Optimization for model identification to increase
RL data efficiency. It identifies a model that is good enough to
predict a policy's value function that is similar to the current
optimal policy. The approach can be used with any policy search
algorithm that guarantees smooth changes in the learned policy. It can
also help the real-world applicability of motion planners that reason
over dynamics and operate with physics engines
\cite{Bekris:2008aa,Littlefield:2017aa}.  Both simulated and
real experiments show that VGMI can decrease the number of rollouts
needed for an optimal policy. Future work includes analyzing the
method's properties, such as expressing the conditions under which the
inclusion of model identification reduces the need for physical
rollouts and the speed-up in convergence. 

\section*{Acknowledgments}
This work was sponsored by NSF IIS-1734492, IIS-1723869, a NASA ECF
award to Dr. Bekris and U.S. Army Research Lab Collaborative Agreement
\# W911NF-10-2-0016.


\bibliographystyle{named}
\bibliography{bib/system_id}

\end{document}